\documentclass[conference]{IEEEtran}
\IEEEoverridecommandlockouts
\usepackage{cite}
\usepackage{amsmath,amssymb,amsfonts}
\usepackage{algorithmic}
\usepackage{graphicx}
\usepackage{textcomp}
\usepackage{xcolor}
\usepackage[most]{tcolorbox}
\usepackage{layouts}
\usepackage{printlen}
\usepackage{pgfplots}
\usepackage{float}
\usepackage{pgfplots}
\usepackage{hyperref}

\usepackage{listings}

\lstdefinestyle{yaml}{
     basicstyle=\color{blue}\footnotesize,
     rulecolor=\color{black},
     string=[s]{'}{'},
     stringstyle=\color{blue},
     comment=[l]{:},
     commentstyle=\color{black},
     morecomment=[l]{-}
 }

\newtcolorbox{myquote}[1][]{%
    colback=black!5,
    colframe=black!5,
    notitle,
    sharp corners,
    borderline west={2pt}{0pt}{black!80!black},
    enhanced,
    breakable,
}

\graphicspath{{figures/}}

\def\BibTeX{{\rm B\kern-.05em{\sc i\kern-.025em b}\kern-.08em
    T\kern-.1667em\lower.7ex\hbox{E}\kern-.125emX}}

\usepackage{todonotes}

\usepackage{csquotes}

\pgfplotsset{compat=1.18}

\begin{document}

\title{Exploring the Carbon Footprint of Hugging Face's ML Models: A Repository Mining Study\\
}


\makeatletter
\renewcommand{\@IEEEsectpunct}{ }
\makeatother

\author{\IEEEauthorblockN{Joel Castaño}
\IEEEauthorblockA{
\textit{\footnotesize Universitat Politècnica de Catalunya}\\
\footnotesize Barcelona, Spain \\
\footnotesize joel.castano@upc.edu}
\and
\IEEEauthorblockN{Silverio Martínez-Fernández}
\IEEEauthorblockA{
\textit{\footnotesize Universitat Politècnica de Catalunya}\\
\footnotesize Barcelona, Spain \\
\footnotesize silverio.martinez@upc.edu}
\and
\IEEEauthorblockN{Xavier Franch}
\IEEEauthorblockA{
\textit{\footnotesize Universitat Politècnica de Catalunya}\\
\footnotesize Barcelona, Spain \\
\footnotesize xavier.franch@upc.edu}
\and
\IEEEauthorblockN{Justus Bogner}
\IEEEauthorblockA{
\textit{\footnotesize University of Stuttgart}\\
\footnotesize Stuttgart, Germany \\
\footnotesize justus.bogner@iste.uni-stuttgart.de}
}


\maketitle



\begin{abstract}
\textit{Background:} The rise of machine learning (ML) systems has exacerbated their carbon footprint due to increased capabilities and model sizes. However, there is scarce knowledge on how the carbon footprint of ML models is actually measured, reported, and evaluated.
\textit{Aims:} This paper analyzes the measurement of the carbon footprint of 1,417 ML models and associated datasets on Hugging Face. Hugging Face is the most popular repository for pretrained ML models. We aim to provide insights and recommendations on how to report and optimize the carbon efficiency of ML models.
\textit{Method:} We conduct the first repository mining study on the Hugging Face Hub API on carbon emissions and answer two research questions: (1) how do ML model creators measure and report carbon emissions on Hugging Face Hub?, and (2) what aspects impact the carbon emissions of training ML models?
\textit{Results:} Key findings from the study include a stalled proportion of carbon emissions-reporting models, a slight decrease in reported carbon footprint on Hugging Face over the past 2 years, and a continued dominance of NLP as the main application domain reporting emissions. The study also uncovers correlations between carbon emissions and various attributes, such as model size, dataset size, ML application domains and performance metrics.
\textit{Conclusions:} The results emphasize the need for software measurements to improve energy reporting practices and the promotion of carbon-efficient model development within the Hugging Face community. To address this issue, we propose two classifications: one for categorizing models based on their carbon emission reporting practices and another for their carbon efficiency. With these classification proposals, we aim to encourage transparency and sustainable model development within the ML community.
\end{abstract}



\section{Introduction}

Sustainability awareness has gained global importance over the last years, driven by various initiatives that emphasize the need for reducing humanity's overall carbon footprint \cite{calero2021introduction}. Given this situation, information and communication technologies (ICTs) have emerged as potential contributors to sustainability, but they can also negatively impact the environment through increased energy consumption, and hence, carbon emissions.
Unfortunately, there is some evidence suggesting that the latter is indeed the case:
Andrae and Edler's projections show the global electricity consumption of ICT increasing from 1,500 TWh (8\% share) in 2010 to up to 30,700 TWh (51\%) in 2030 \cite{challe6010117}. This underlines the importance of sustainability in IT research to minimize the negative impact of ICT on the environment \cite{penzenstadler2012sustainability}.

As a subset of ICTs, systems incorporating machine learning (ML), i.e., ML-based systems~\cite{Martinez-Fernandez2022}, are especially popular lately. Nevertheless, this popularity in ML capabilities has come with an increase in model size and training time \cite{schwartz2020green}. Unfortunately, a substantial increase in the energy consumption of ML models could lead to significant environmental impact. Thus, it is crucial to develop inclusive and environmentally friendly ML-based systems. This aligns with global efforts to reduce carbon emissions in all sectors of society.

While several recent studies shed light on how energy efficiency can be increased during model training (e.g., during hyperparameter tuning \cite{stamoulis2018hyperpower}), little is known about the actual emissions of most published ML models, or even if creators of ML models consider and report energy-related aspects when publishing their models.
Potential sources for this information are public repositories of pretrained models.
One well-known example of such repositories is the Hugging Face Hub~\cite{HuggingFaceInc.2023}, which has emerged as the most popular platform for pretrained language models and associated datasets~\cite{Jain2022}, and also for computer vision or audio processing models. In spite of this widespread use, there is currently a lack of understanding how carbon emissions are reported in this repository.

In this context, our study investigates the reported carbon emissions of ML models during training, and how they compare in terms of carbon efficiency. We chose repository mining \cite{repositoryMiningStandard} as it allows us to quantitatively analyze the large-scale Hugging Face dataset of models and their carbon emissions, providing insights into carbon efficiency patterns in ML models. Hugging Face is a suitable data source due to its extensive collection of pretrained models, popularity among ML practitioners, and the availability of metadata. Moreover, the Hugging Face Hub API and HfApi class enable us to systematically extract information from the repository. By analyzing the carbon emissions of various ML models on Hugging Face, we aim to provide insights and recommendations for ML practitioners and researchers who are looking to report and optimize the carbon efficiency of their models. 

\textbf{Data availability statement}: Our replication package, including the datasets, code, and detailed documentation, is available on Zenodo \cite{castano_fernandez_joel_2023_8115187}. 
It contains Jupyter notebooks with the data extraction, preprocessing, and analysis scripts, along with the raw, processed, and manually curated datasets used for the analysis. We have provided explanations on how to navigate the data source, how to use it, and how the provided data, code, and tools are used in the steps of the methodology described in the paper.

\section{Background and Related Work}
In this section, we explain the most important concepts to understand this study and present related work in the area.

\subsection{Sustainability and Energy Consumption of Software}
In general, sustainability represents \enquote{the capacity of something to last a long time}~\cite{calero2021introduction}, but is also associated with \enquote{the resources used}~\cite{calero2021introduction} for a particular activity.
Regarding software sustainability, the Karlskrona Manifesto~\cite{Becker2015} describes five dimensions: technical, economic, social, individual, and environmental.
Our study exclusively targets the latter dimension.
The movements to increase energy efficiency and reduce the carbon footprint of software have been called \textit{Green IT} and \textit{Green Software}~\cite{Verdecchia2021a}, and the complementary concept for artificial intelligence is \textit{Green AI}~\cite{schwartz2020green}~\cite{Georgiou2022}.
Within Green AI, the training of ML models is especially important because its long duration and extensive use of computing hardware can make it especially resource-intensive~\cite{Strubell2019}.

One of the most important aspects of environmental software sustainability is \textit{energy consumption}, which is usually measured in \textit{joule} (J) or \textit{kilowatt-hours} (kWh)~\cite{Cruz2021}.
In this sense, we may want to know how much energy a certain operation consumed, e.g., the training of an ML model.
Most tools usually measure and report the average energy or power consumption for a certain timeframe~\cite{Cruz2022}.

While measuring energy consumption is useful to judge energy efficiency, the extent of harmful greenhouse gases like carbon dioxide (CO$_2$) depends on how the electrical energy was produced.
Therefore, a more accurate concept to measure climate impact is the \textit{carbon dioxide equivalent} (CO$_2$e), colloquially known as the \textit{carbon footprint}~\cite{Cruz2021}.
It is reported in mass units, e.g., kgCO$_2$e or gCO$_2$e, and is the suggested metric in the greenhouse gases standard ISO 14064~\cite{InternationalOrganizationForStandardization2018}.
A model trained in an Icelandic data center with geothermally produced electricity may still consume substantial energy, but can have a marginal carbon footprint.
While many more tools exist to measure software energy consumption, libraries like \texttt{CodeCarbon}~\cite{CodeCarbon} or Lacoste et al.'s ML workload calculator~\cite{lacoste2019quantifying} can estimate the carbon footprint.

\subsection{Hugging Face Hub}
Since training complex ML models requires considerable expertise and resources, it is desirable to reuse pretrained models.
The company Hugging Face, Inc. provides a community platform to achieve this.
Originally founded in 2016 as an NLP company, Hugging Face gained popularity for open-sourcing their NLP models~\cite{Jain2022} and for creating an easy-to-use Python library for NLP transformers~\cite{wolf2019huggingface}.
Today, one of their most important offers is the Hugging Face Hub, which is an open platform to publish ML models and datasets.
For documentation and reproducibility, the Hub adopted Mitchell et al.'s \textit{Model Cards} idea~\cite{mitchell2019model}.
Each published model can provide a \texttt{README.md} plus additional metadata, e.g., performance metrics or tags.
This also includes a carbon footprint attribute (\textit{co2\_eq\_emissions}), with guidelines on how to report it~\cite{huggingfaceDisplayingCarbon}.
Lastly, a convenient option to train and publish models is to use the \textit{AutoTrain} feature~\cite{HuggingFaceAutoTrain}, which is the AutoML infrastructure of Hugging Face.
Models published via AutoTrain automatically include their training carbon footprint.

\subsection{Related Work}
Since Schwartz et al.'s seminal paper on Red vs. Green AI in 2020~\cite{schwartz2020green}, research in this area has increased steadily.
A recent literature review on Green AI by Verdecchia et al.~\cite{verdecchia2023systematic} identified 98 primary studies, with the most prevalent topics being energy monitoring of ML models throughout their lifecycle (28 papers), energy efficiency of hyperparameter tuning (18), and energy footprint benchmarking of different ML models (17).
From a more foundational perspective, García-Martín et al.~\cite{Garcia-Martin2019} synthesized guidelines and tools to estimate the energy consumption of ML models.
Similarly, Patterson et al.~\cite{Patterson2022} proposed four general practices to reduce the carbon footprint of ML training, namely the 4Ms: \textit{model}: selecting efficient model architectures, e.g., sparse models; \textit{machine}: using processors optimized for ML training, e.g., tensor processing units (TPUs); \textit{mechanization}: using optimized cloud data centers; and \textit{map}: picking a data center location with clean energy.

Several studies also conducted energy consumption benchmarks of various model characteristics or training methods, especially for deep learning.
Yarally et al.~\cite{yarally2023uncovering} compared different hyperparameter optimization methods and layer types, while Xu et al.~\cite{Xu2023} analyzed the impact of network architectures, training location, and measurement tools on both energy consumption and carbon footprint. Santiago del Rey et al.~\cite{delrey2023dl} analyzed the impact of the model architecture and training environment when training computer vision models.
Verdecchia et al.~\cite{Verdecchia2022} analyzed how exclusively modifying the dataset can reduce energy consumption, and Brownlee et al.~\cite{Brownlee2021} conducted one of the few studies that analyzed the tradeoff between accuracy and energy consumption during hyperparameter optimization not only for model training, but also for the inference stage.
Regarding ML frameworks, Gergiou et al.~\cite{Georgiou2022} performed an in-depth comparison of the energy consumption between TensorFlow and PyTorch. With respect to ML domains (NLP, vision, etc.) and its carbon emissions, Bannour et al.~\cite{bannour-etal-2021-evaluating} evaluated the carbon footprint of NLP methods and its measuring tools. Lastly, in the context of mobile applications containing neural networks, Creus et al.~\cite{castanyer2021design} published a registered report at ESEM'21 about identifying design decisions with impact on energy consumption.

However, to the best of our knowledge, no study has so far analyzed the carbon footprint of training ML models from a large-scale repository mining perspective.
While the many benchmark-based studies provide guidance on how to improve energy efficiency, our study reports insights into the current state of carbon emission reporting in one of the largest ML communities.
Closely related to our endeavor is an interview study by Jiang et al.~\cite{jiang2023empirical}: they synthesized practices and challenges for the reuse of pretrained models in the Hugging Face ecosystem, and complemented their interviews with a security risk analysis enriched with data queried from the Hugging Face Hub.
However, energy consumption was not part of their study, and also not mentioned by their interviewees.

\section{Methodology}

\begin{figure}[!htbp]
    \centering
    \includegraphics[width=0.9\columnwidth]{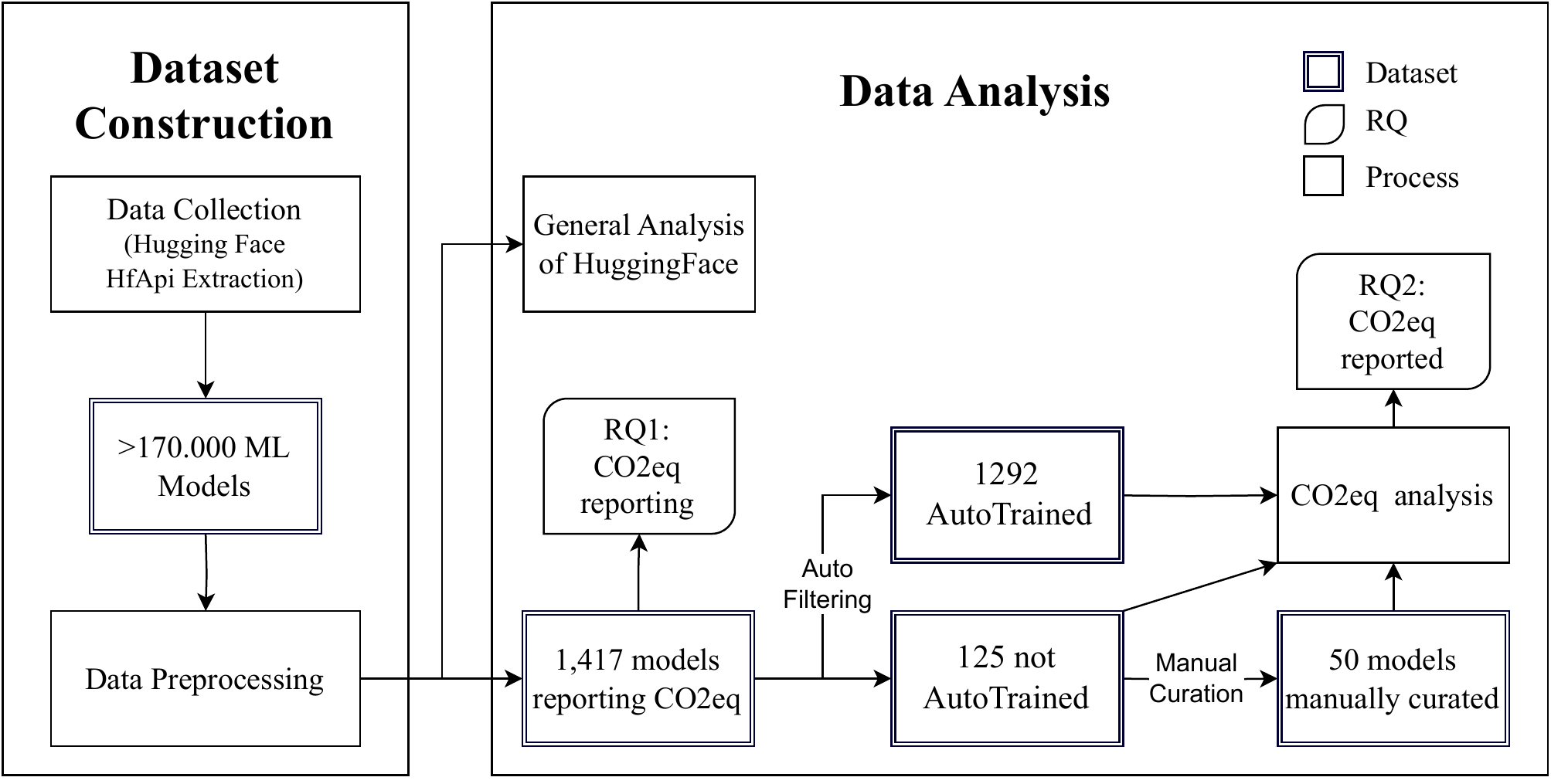}
    \caption{Data collection and analysis process during the study}
    \label{study design evolution}
\end{figure}

In this section, we first define the study objective and the research questions that will guide the investigation. Next, we explain how to obtain the dataset for the analysis of the research questions. Figure~\ref{study design evolution} illustrates the outline of the data collection procedure along with the design of our investigation.

\subsection{Study Objective and Research Questions}
We formulate our research goal according to the Goal Question Metric (GQM) guidelines \cite{caldiera1994goal} as follows:\\
Analyze \textit{ML models and corresponding datasets} with the purpose of \textit{understanding and measuring} with respect to \textit{carbon emissions during training} from the point of view of \textit{the ML practitioner} in the context of \textit{the Hugging Face Hub}.

This goal breaks into two main Research Questions (RQ), which respectively have several sub research questions. Firstly, we seek to understand the practices surrounding the measurement and reporting of training carbon emissions for ML models on Hugging Face:

\begin{myquote}
\textbf{RQ1}. \textit{How do ML model creators measure and report training carbon emissions on Hugging Face Hub?}
\end{myquote}

\begin{itemize}
    \item RQ1.1: How has the reporting of carbon emissions evolved over the years?
    \item RQ1.2: How has the reported carbon emissions evolved over the years?
    \item RQ1.3: What are the main characteristics of the models reporting carbon emissions?
    \item RQ1.4: How can we classify Hugging Face models based on their carbon emission reporting?
\end{itemize}

Subsequently, we conduct an in-depth analysis of various factors related to the carbon emissions of training ML models. We aim to identify and comprehend the diverse elements that influence the carbon emissions during the training process:

\begin{myquote}
\textbf{RQ2}. \textit{What aspects impact the carbon emissions of training ML models on Hugging Face Hub?}
\end{myquote}

\begin{itemize}
    \item RQ2.1: How are carbon emissions and model performance related?
    \item RQ2.2: How are carbon emissions related to model and dataset size?
    \item RQ2.3: What is the difference in carbon emissions between fine-tuned and pretrained tasks?
    \item RQ2.4: How do ML application domains affect carbon emissions?
    \item RQ2.5: How can we classify Hugging Face based on their carbon efficiency?
\end{itemize}

\subsection{Dataset Construction}

In our repository mining study \cite{repositoryMiningStandard}, we start by gathering and preprocessing the data to answer the above RQs.

\subsubsection{\textbf{Data Collection}}

We developed a data collection pipeline as part of the replication package to collect data from the Hugging Face models and their associated Model Cards. The pipeline uses the Hugging Face Hub API through the HfApi class, a Python wrapper for the API, which provides a \emph{list\_models} function. This function allows us to retrieve all the information of every model that has been uploaded to Hugging Face and to apply subset filtering. In this study, we collected data up to March 2023 (included). The replication package \cite{castano_fernandez_joel_2023_8115187} allows updating the data in the future.

Once we have the models list retrieved, the pipeline automates the extraction of information, including the use of regular expressions to locate and collect evaluation metrics such as accuracy, F1, Rouge1, and so on. Each row of the dataframe generated by the pipeline is a model alongside many attributes and its CO$_2$e emissions metric if it is reported. Initially, we collected every model of Hugging Face regardless of whether carbon emission was reported or not. Afterwards, we consider the dataset split based on the models that reported CO$_2$e. The main attributes that we collected are the following:

\begin{itemize}
    \item \textit{co2\_eq\_emissions}: the reported training CO$_2$e emissions.
    \item \textit{datasets\_size}: the total size of the datasets used. 
    \item \textit{training\_type}: if pretraining or fine-tuning training.
    \item \textit{geographical\_location}: the location of the training.
    \item \textit{hardware\_used}: the hardware used for the training.
    \item \textit{performance metrics}: evaluation metrics (accuracy, F1, Rouge1, RougeL) if reported.
    \item \textit{size}: size of the final model trained.
    \item \textit{auto}: if the model is AutoTrained.
    \item \textit{downloads}: number of downloads of the model.
    \item \textit{library\_name}: library used by the model.
    \item \textit{created\_at}: date of the model creation on Hugging Face
    \item \textit{tags}: the attached tags (e.g., PyTorch, Transformer, \ldots)
    
\end{itemize}

Despite its ability to handle various data formats and filter out unnecessary information, the pipeline cannot fully address the issues of missing values and varying reporting formats due to the lack of a strict standard for information reporting on Hugging Face models.

For attributes like \textit{datasets\_size}, where manual data collection is required, we have added the information manually to the dataset. We acknowledge that this manual data collection process may not be entirely reproducible and could introduce some inconsistencies.

\subsubsection{\textbf{Data Preprocessing}}

After gathering all the data, we end up with a dataframe with all the models of Hugging Face, in particular 170,464 data entries. Only 1,417 of these models report the carbon emissions needed to train the model. Further, we clean this dataset to homogenize the diverse data formats, enabling easier analysis. We followed these steps in our data preprocessing: a) feature engineering for further analyses; b) variable harmonization, curation, and filtering (e.g., CO$_2$e units harmonization); c) one-hot encoding of tags; d) filtering of tags by deleting language and auxiliary tags (e.g., 'arxiv', 'doi', etc.) and tags that are included in less than 100 models; and e) creating a dictionary relating tags to their ML application domain.

We focus on feature engineering, variable standardization, and one-hot encoding of tags. We create variables such as \textit{co2\_reported}, \textit{auto}, \textit{year\_month}, and \textit{domain} for filtering, splitting datasets, and analyzing model behavior across ML application domains. The domain variable was extracted through a mapping on the model tags to their corresponding domains, including: Multimodal, Computer Vision, NLP, Audio, and Reinforcement Learning.

We standardize variables like CO$_2$e, which are reported in different units (e.g., kg or g) or the \textit{created\_at} variable by converting it to datetime format. Also, we filter out models that may have inaccurate CO$_2$e reporting (e.g., CO$_2$e = 0).

Lastly, we one-hot encode the tags variable, which we gathered as a list in the tags attribute for each model during data collection. We then filter the one-hot encoded tags, removing irrelevant ones such as language tags or auxiliary tags like 'license', 'dataset', or 'doi'.

\subsubsection{\textbf{Manually curated carbon emission dataset and final datasets}}

After data preprocessing, our dataset contains over 170,000 models, with 1,417 reporting carbon emissions. Approximately 1,293 of the models reporting carbon emissions are AutoTrained. Many of the AutoTrained models are suspected to be of low quality, as they do not report any attribute nor text documentation along the emissions. Moreover, some attributes cannot be automatically retrieved if not reported in the model card. Therefore, we manually improve the metadata of around 150 non-AutoTrained models by completing missing attributes, including performance metrics, filtering out models without documentation or with suspicious reporting, and removing excessively repeated baseline models fine-tuned for different tasks.
This results in a cleaner, manually curated carbon emission dataset of around 50 models, used alongside the complete Hugging Face dataset (170,464 models) for general context analysis and the carbon emission dataset (1,417 models) for addressing RQ1 and RQ2.

\subsection{Data Analysis}

Next, we explain the data analysis methodology for the reproducibility of the results section.

\subsubsection{\textbf{RQ1 Analysis}} We use the complete carbon emission dataset.
To answer how the reporting of carbon emission evolves (RQ1.1), we analyze the monthly percentage of models reporting carbon emissions. To conclude with statistical significance if a trend exists, we perform a t-test on the slope parameters of a linear regression fitted to this evolution, with the null hypothesis ($H_0$) that there is no linear trend. We check the linear regression assumptions (residuals normality and homoscedasticity, among others) to ensure the accuracy of the t-test. In light of multiple hypotheses testing throughout the study, we apply the Holm-Bonferroni correction \cite{abdi2010holm} to adjust the p-values, ensuring accurate comparisons to our fixed significance level ($\alpha$ = 0.05). This approach helps control the familywise error rate and reduces the risk of false positives.

To study the evolution of the carbon emissions (RQ1.2), we perform a similar analysis with the monthly trend of the median \textit{co2\_eq\_emissions}. We use the median rather than the mean, as there are outlier models reporting extreme carbon emissions that distort the analysis, e.g., recent large language models (LLMs). Equivalently, we perform a t-test on the slope of a fitted linear regression.

To study the evolution of the ML application domain trends on Hugging Face (RQ1.3), we use the harmonized domain variable along with other descriptive attributes.

Finally, for RQ1.4 classification criteria on carbon emissions reported, we develop a classification system that evaluates models based on their carbon emission reporting practices. This classification aims to provide a clearer understanding of the Hugging Face carbon emission reporting, encouraging more transparent reporting.


\subsubsection{\textbf{RQ2 Analysis}} We use the complete carbon emission dataset and the cleaned non-AutoTrained models dataset.

We perform a correlation analysis on several factors. The selection of these factors is based on the available data and the assumption that these could be related to the resource usage, and consequently, the carbon consumption of ML models. Dataset and model size are chosen as there is evidence on the role of these attributes regarding energy consumption \cite{Verdecchia2022}. The performance metrics are included as we aim to understand if there is any trade-off between model efficiency and energy consumption, which can guide the development of more energy-efficient yet powerful models. Further, the types of tasks (like pretraining or fine-tuning) and application domains (like NLP or Computer Vision) are chosen because they represent different computational workloads and patterns that can impact energy consumption (some studies, such as Bannour et al.~\cite{bannour-etal-2021-evaluating}, already try to assess energy consumption by domain). Other potentially interesting comparisons, such as local vs cloud training, are not included due to the lack of available data.

First, we evaluate if there is a trade-off between carbon emissions and model performance (RQ2.1). We calculate Spearman's correlation coefficient between each performance metric in the dataset (accuracy, F1, Rouge1, and RougeL) and reported CO$_2$e emissions, for both AutoTrained and manually curated non-AutoTrained models. We consider Spearman's non-parametric correlation rather than Pearson's as the performance metrics are not normally distributed.
A significant positive correlation would indicate that this trade-off exists.

Next, we examine whether larger models and datasets are associated with increased carbon emissions (RQ2.2). To investigate this, we perform a Pearson correlation test applying log-transformations on the variables. The log transformation is applied to ensure the normality of the variables so that Pearson can be applied, as the three variables exhibit power law distributions. We also considered the manually curated dataset, as the \textit{datasets\_size} variable was manually extracted.

To study if fine-tuning tasks are more carbon-efficient than pretraining tasks (RQ2.3), we perform a Mann-Whitney U test at significance level $\alpha=0.05$. As the reported CO$_2$e does not follow a normal distribution (the Shapiro-Wilk test shows a p-value of $\approx$ 0, with heavy tails and kurtosis), we use a non-parametric test comparing the ranks of the data points in the two groups, rather than their means via a t-test. 


To study if some ML application domains are more carbon-efficient than others (RQ2.4), we again perform a Mann-Whitney U test with significance level $\alpha=0.05$.

Finally, for the classification criteria study (RQ2.5), we have developed a carbon efficiency classification system. Carbon efficiency is the ability to minimize greenhouse gas emissions (e.g, CO2), per unit of output or service provided. In the context of ML models, we will refer as carbon efficiency to the model capacity of minimizing emissions generated during training and deployment processes.
We have adopted the use of index values as proposed in \cite{europeanCommission} and used in \cite{fischer2023unified}. Indexing allows us to put all attribute values in relation to reference values, effectively dropping the units of the attributes and enabling easier comparisons. The carbon efficiency classification system used in this study is based on the weighted mean of the quartiles of the model for each indexed attribute. We consider the following attributes:

\begin{itemize}
	\item CO$_2$e emissions: the caused carbon footprint.
	\item Size efficiency: the ratio of the model size to CO$_2$e  emissions, which favors models that achieve low emissions even with high model complexity.
    \item Dataset efficiency: the ratio of the datasets size to CO$_2$e emissions, which favors models that achieve low emissions even with large datasets.
	\item Downloads: a proxy for reusability, with more downloads suggesting greater efficiency through reuse.
	\item Performance score: the harmonic mean of normalized performance metrics (accuracy, F1, Rouge-1, Rouge-L), penalizing cases where a single metric is extremely low.
\end{itemize}

The attributes are assigned weights based on their significance for carbon efficiency: CO$_2$e emissions (0.35), model size efficiency (0.1), datasets size efficiency (0.1), downloads (0.25), and performance score (0.2). \texttt{distilgpt2} serves as the reference model for index values due to its high download count among models that report carbon footprint. The index reference is structured so that lower index values indicate better efficiency for a given attribute. Consequently, for attributes that should be maximized (size efficiency, downloads, and performance), the index is calculated as $i = \text{ref} / \text{val}$; for attributes that should be minimized (CO$_2$e emissions), $i = \text{val} / \text{ref}$. This allows us to classify each of the carbon emissions-reporting models into five carbon efficiency labels, from E to A (from less to more carbon efficient).

\section{Results}

In this section, we present a general analysis of the extracted Hugging Face model dataset, and then the results per RQ.

\subsection{General Analysis of Hugging Face Models}

As a preliminary analysis, we report general characteristics of the ML models stored in the Hugging Face repository.

\subsubsection{\textbf{Is Hugging Face's popularity increasing?}}
\begin{figure}[H]
    \centering
    \includegraphics[width=0.85\columnwidth]{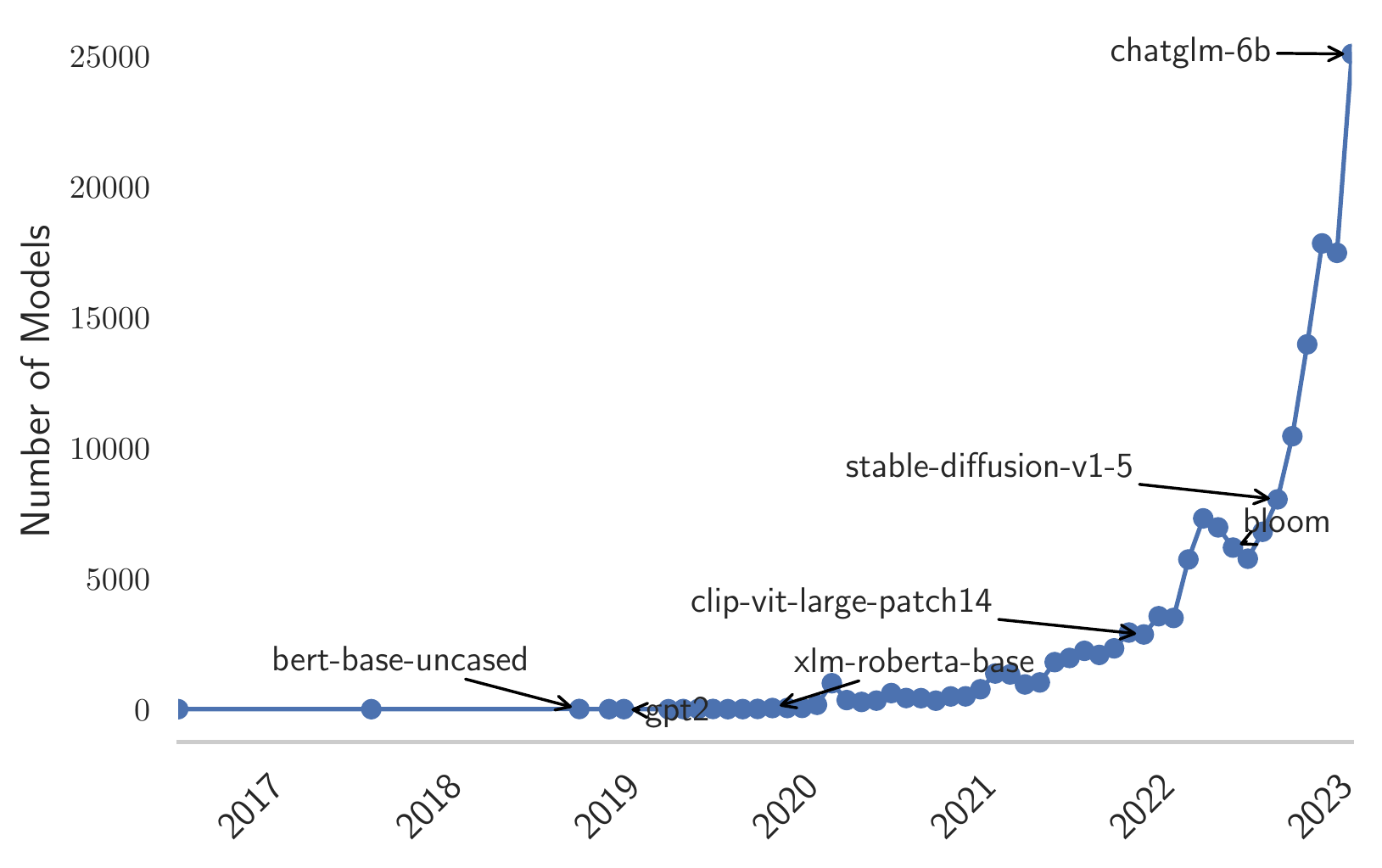}
    \caption{Evolution of total number of Hugging Face models added each month}
    \label{Hugging Face_evolution}
\end{figure}

From the number of models uploaded each month in Figure~\ref{Hugging Face_evolution}, we can observe that Hugging Face's popularity has been increasing exponentially in the past years.

\subsubsection{\textbf{What is the main ML application domain in Hugging Face?}}

\begin{figure}[!htbp]
    \centering
    \includegraphics[width=0.85\columnwidth]{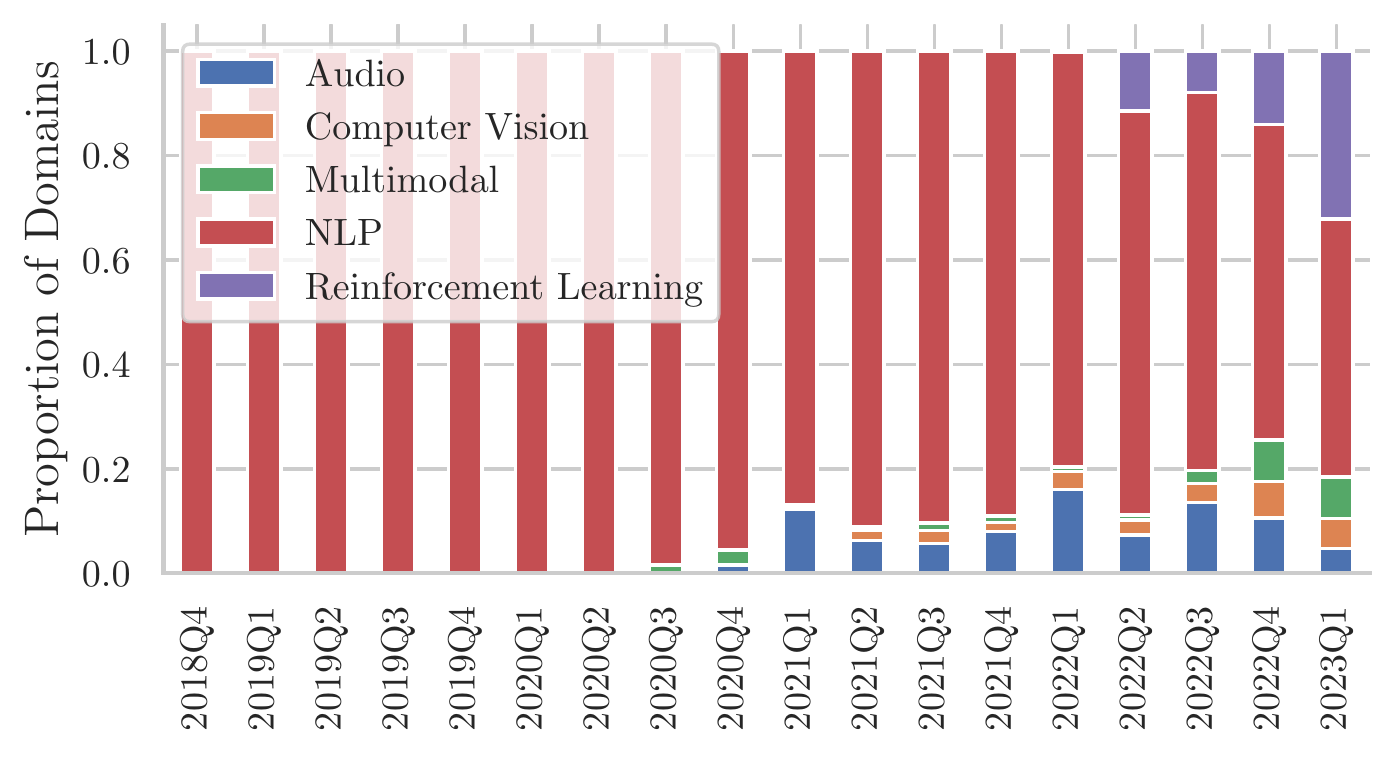}
    \caption{Application domain evolution of Hugging Face models}
    \label{model_domains_evolution}
\end{figure}

Hugging Face models are tagged by users with information on various characteristics such as the model's task, the library used, and other configuration-related tags. The most popular tag is the \textit{transformer} tag, along with \textit{PyTorch}, the most used library in Hugging Face. The majority of the remaining tags are NLP-related, together with auxiliary tags (e.g., \textit{autotrain\_compatible}) or tags related to reinforcement learning. This NLP dominance can be further seen in Figure~\ref{model_domains_evolution}. NLP has been the dominant ML application domain since Hugging Face started. In fact, NLP was the only domain until the end of 2020, when multimodal or audio models started to appear. In the past months, the NLP dominance has been shrinking more than ever, with an increase in reinforcement learning models and other domains. Lastly, almost 30\% of models have been published without any tag (\texttt{no-tag}), making their categorization impossible.

\subsection{Carbon Emissions Reporting (RQ1)}

\subsubsection{\textbf{How has the reporting of carbon emissions evolved over the years?}}

~

\begin{myquote}
\textbf{Finding 1.1}. \textit{Despite more models reporting carbon emissions, the percentage they represent over all models published in Hugging Face is not just marginal but stalled, pointing out lack of awareness of green AI by the community.}
\end{myquote}

\begin{figure}[h]
    \centering
    \includegraphics[width=0.85\columnwidth]{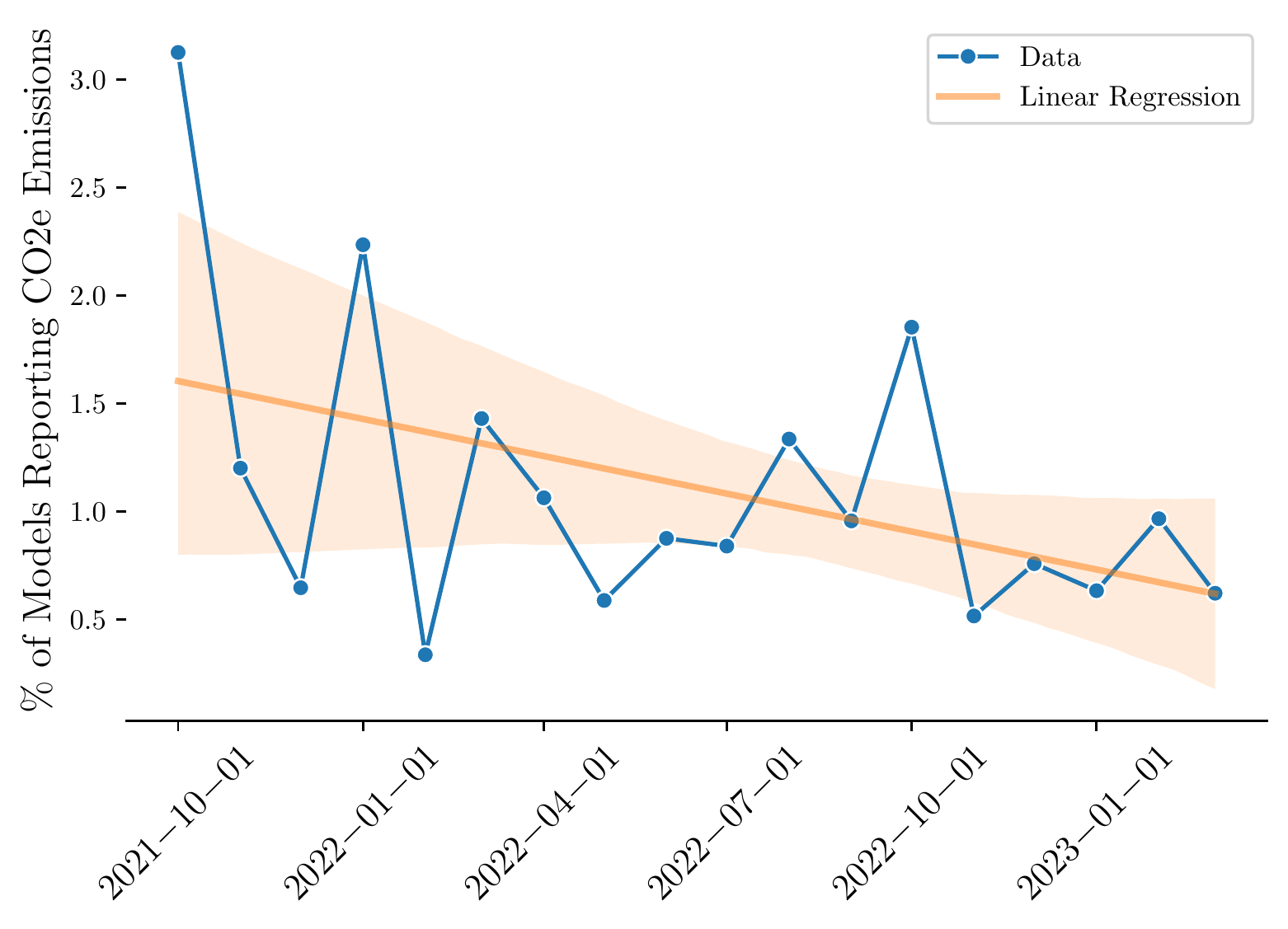}
    \caption{Evolution of the carbon emission reporting on Hugging Face}
    \label{energy_consumption_evolution}
\end{figure}

As illustrated in Figure~\ref{Hugging Face_evolution}, the popularity of Hugging Face has increased significantly, with a growing number of models being published each month. Consequently, the number of models reporting their carbon emissions has also risen. However, to understand the overall interest in reporting carbon emissions or awareness of Green AI, we need to examine the evolution of the percentage of models that report their carbon emissions. Figure~\ref{energy_consumption_evolution} presents the monthly aggregates for this.

Carbon emission reporting on Hugging Face started in mid-2021. However, this feature has unfortunately not experimented wide-spread adoption so far. The maximum percentage occurred in 10-2021, with 3.12\% of the models reporting carbon emissions. The average percentage is 0.9\%, and for the last month (03-2023), it is 0.62\%. The median percentage is 0.92\%. With the t-test result on the regression slopes (adjusted p-value $=0.57$), we do not have enough evidence to consider a decreasing trend. However, the stall may indicate a lack of awareness or no perceived importance of Green AI concerns in the Hugging Face community.

\subsubsection{\textbf{How have the reported carbon emissions evolved over the years?}}
~
\begin{myquote}
\textbf{Finding 1.2}. \textit{The carbon emissions reported on Hugging Face have slightly decreased in the past 2 years.}
\end{myquote}

\begin{figure}[h]
    \centering
    \includegraphics[width=0.85\columnwidth]{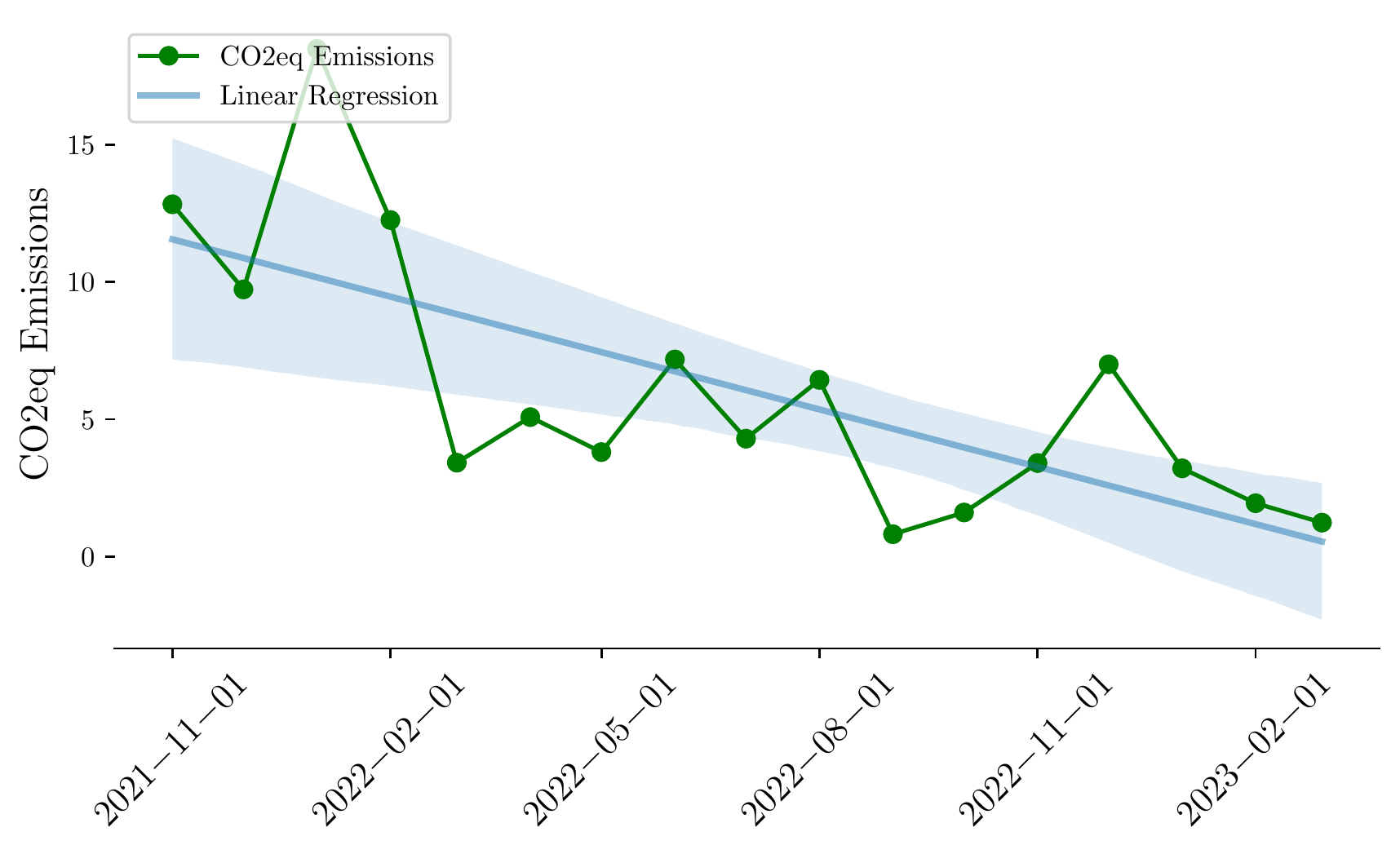}
    \caption{Evolution of the reported carbon emissions on Hugging Face}
    \label{evolution_energy_reported}
\end{figure}

In addition to analyzing the practice of carbon emission reporting, we also wanted to know if the reported carbon emissions increased, i.e., if recent models are more demanding. We can observe the evolution of the median carbon emissions aggregated by month in Figure~\ref{evolution_energy_reported}. 10-2021 was not included in the figure, as it was the first month in which carbon emissions were published on Hugging Face, with an abnormally high median CO$_2$e (65.58g) compared to other months.

Contrary to expectations, the median reported CO$_2$e has been slightly decreasing over the past months. Excluding 10-2021, the maximum was reached in 01-2022 with 18.48g. For the whole timeframe, the median was 4.69g and the mean was 9.35g. The latter is roughly equivalent to the generated emissions by fully charging a single phone. In the last included month (03-2023), the reported emissions dropped to 1.24g. Based on the slope t-test (adjusted p-value of $0.011$), we can conclude that there has been a statistically significant decrease in the reported carbon emissions over the last 2 years. The observed decrease might indicate a growing awareness and adoption of energy-efficient practices among AI researchers. Additionally, this trend could initially have been influenced by the carbon emissions reports of popular, yet energy-intensive models. As more diverse practitioners (including those at an amateur level) started to report carbon emissions, the reported carbon consumption could have been driven down due to potentially less complex and less carbon-consuming models.

\subsubsection{\textbf{What are the main characteristics of the models reporting their carbon emissions?}}
~
\begin{myquote}
\textbf{Finding 1.3}. \textit{While NLP dominates the carbon emissions reporting, computer vision shows the highest proportion of models within its domain reporting emissions in recent quarters. Other domains remain marginal.}
\end{myquote}

\begin{figure}[h]
    \centering
    \includegraphics[width=0.8\columnwidth]{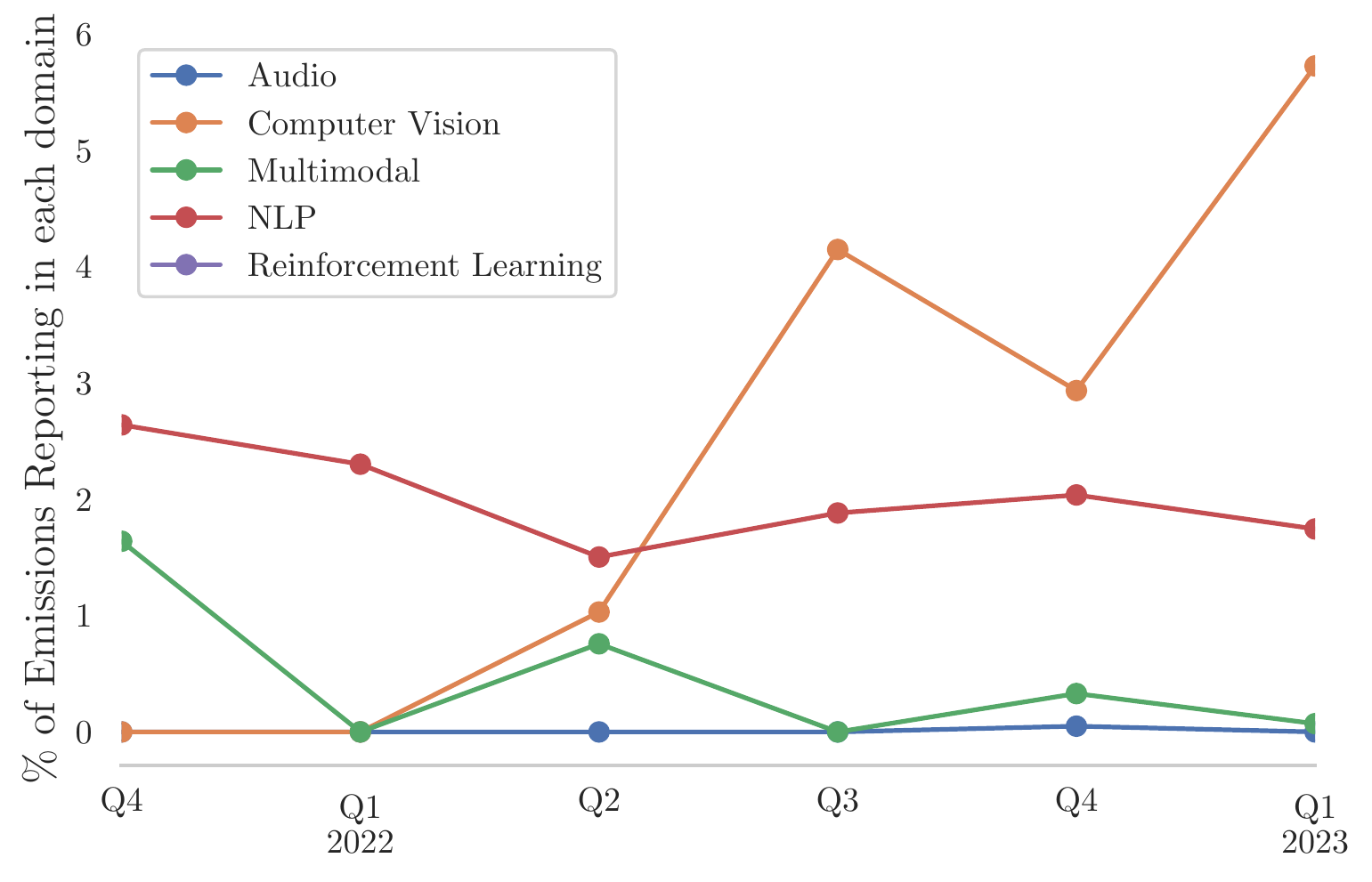}
    \caption{Evolution of the \% of emissions reporting models in each domain}
    \label{domain_evolution_energy}
\end{figure}

As we saw in Figure \ref{model_domains_evolution}, NLP has been the main domain in Hugging Face for the past years. As expected, this domination is replicated for those models reporting carbon emissions ($\approx85\%$ of the carbon emissions models are NLP). Additionally, in Figure~\ref{domain_evolution_energy} we can visualize the evolution of the percentage of models in each domain reporting carbon emissions. Computer vision seems to be the domain showing the largest relative percentage of models reporting carbon emissions (e.g., for the last quarter, roughly 6\% of computer vision models reported carbon emissions). Moreover, the other domains apart from NLP and computer vision seem to be rather marginal.

As a final remark regarding the characteristics of emissions models, over 90\% of the models that report carbon emissions are AutoTrained, and only AutoTrained models provide performance metrics. AutoTrained models automatically report their performance metrics and carbon emissions upon publication. Consequently, many users might not have consciously reported this information, but rather did so as a result of the AutoTrain process. Furthermore, by examining their tags, we can see that 95\% of the emissions reporting models use PyTorch, and at least 78\% use the Hugging Face Transformers library. Therefore, it is safe to assume that most of the algorithms involved in these models are variants of neural networks. Unfortunately, the lack of data for a more detailed examination of the algorithms used prevents us from considering the algorithm efficiency on further analyses.

\subsubsection{\textbf{How can we classify Hugging Face models based on their carbon emission reporting?}}

\begin{figure}[h]
    \centering
    \includegraphics[width=0.85\columnwidth]{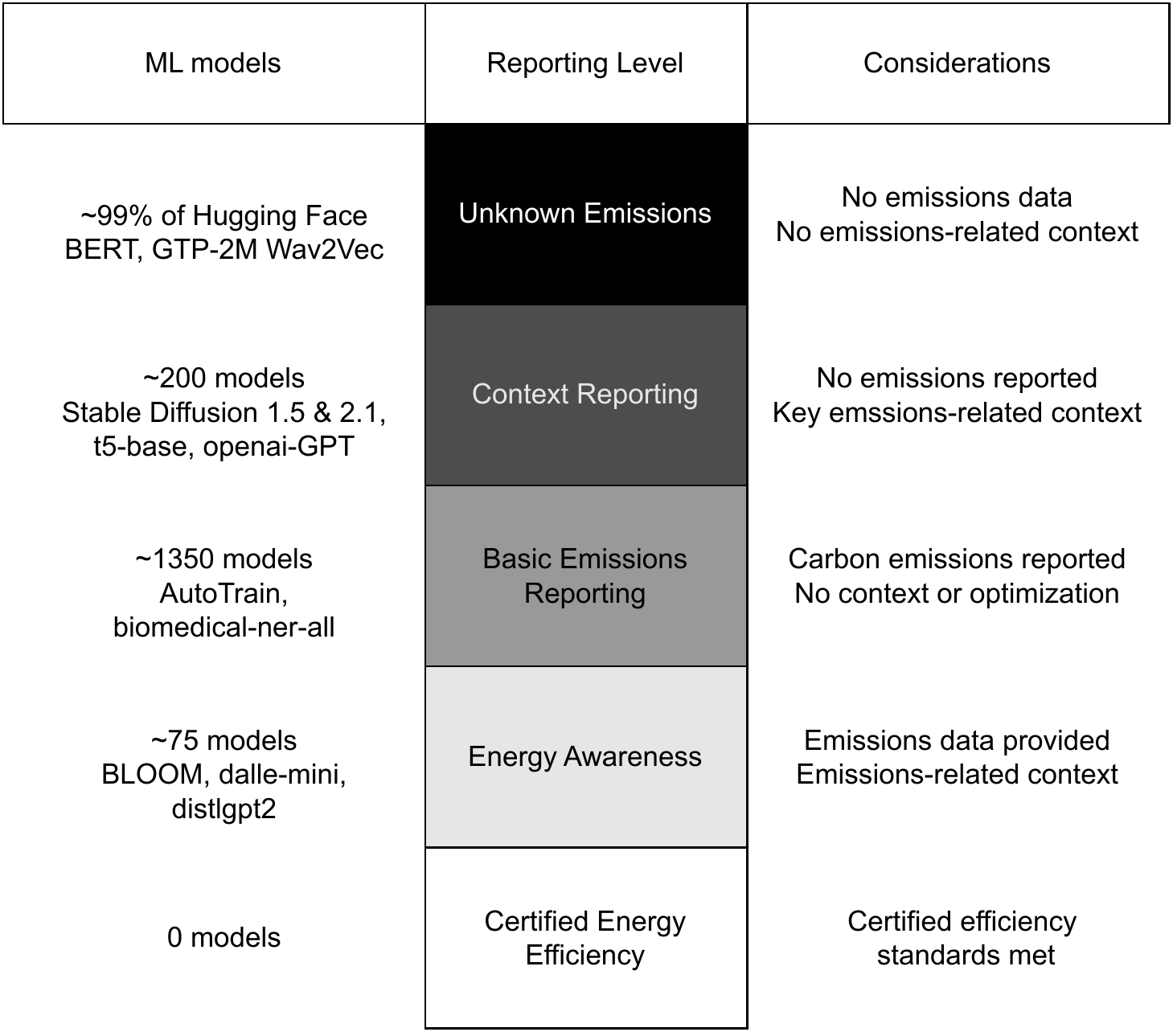}
    \caption{Carbon emission reporting classification}
    \label{Energy reporting classification}
\end{figure}

Using our classification scheme, our analysis reveals a significant disparity in the carbon emission reporting practices present on the platform. The results can be found in Figure~\ref{Energy reporting classification}

\begin{enumerate}
\item Unknown Emissions (99\% of models): The vast majority of Hugging Face, including well-known models such as \texttt{BERT}, \texttt{GPT-2}, and \texttt{Wav2Vec}, do not report any emissions-related information (neither emissions nor context), which hides their environmental impact and inhibits the development of carbon-efficient practices.
\item Context Reporting ($\approx$200 models): This category includes models that do not provide emission data but report some contextual attributes related to carbon emissions, e.g., the hardware used, the dataset size, or training location. Some notable models in this group are \texttt{Stable Diffusion 1.5 \& 2.1}, \texttt{t5-base}, and \texttt{openai-GPT}. While explicit emission data is preferable, the provided training context is a step forward towards estimating energy consumption.
\item Basic Emission Reporting ($\approx$1,350 models): Models in this category report carbon emission data without context or optimization efforts. This group includes most of the AutoTrained models, \texttt{biomedical-ner-all}, and others. While providing carbon emission data is an improvement over not reporting it, the lack of context and optimization limits the usefulness.
\item Energy Awareness ($\approx$75 models): These models report both carbon emissions and context, indicating an increased awareness of carbon efficiency concerns. Examples are \texttt{BLOOM}, \texttt{dalle-mini}, and \texttt{distilgpt2}. Sharing both carbon emission data and context contributes to a better understanding of the relationship between carbon emissions and model performance.
\item Certified Energy Efficiency (0 models): Unfortunately, no models on Hugging Face currently fall into this category, which requires optimized carbon emissions and adherence to established energy efficiency categorizations such as the ones proposed in \cite{fischer2023unified} or \cite{deneckere2020ecosoft}.
\end{enumerate}

\subsection{Correlations Between Carbon Emissions and Other Attributes (RQ2)}

\subsubsection{\textbf{How are carbon emissions and model performance related?}}

~

\begin{myquote}
\textbf{Finding 2.1}. \textit{We could not find a correlation between model performance and carbon emissions for Hugging Face models reporting carbon emissions.}
\end{myquote}

Since only AutoTrained models report performance metrics for the carbon emissions models, we only considered these models in this analysis. Using Spearman correlation, we obtain correlations close to 0 with p-values $>0.05$ for every performance metric. Thus, it is reasonable to assume that this trade-off does not exist for AutoTrained models on Hugging Face. This result may be due to the nature of the metrics reported by the AutoTrain method, as they might represent non-reliable results. To further investigate this, we consider the manually curated dataset ($n=48$). However, with p-values $>0.05$ and correlations between -0.3 and 0.3, we again conclude that there is no evidence in our sample to support that carbon emissions and model performance metrics are related.

\subsubsection{\textbf{How are carbon emissions related to model and dataset size?}}

~

\begin{myquote}
\textbf{Finding 2.2}. \textit{Larger models and larger datasets imply an increase in carbon emissions during training.}
\end{myquote}

\begin{figure}[h]
    \centering
    \includegraphics[width=0.85\columnwidth]{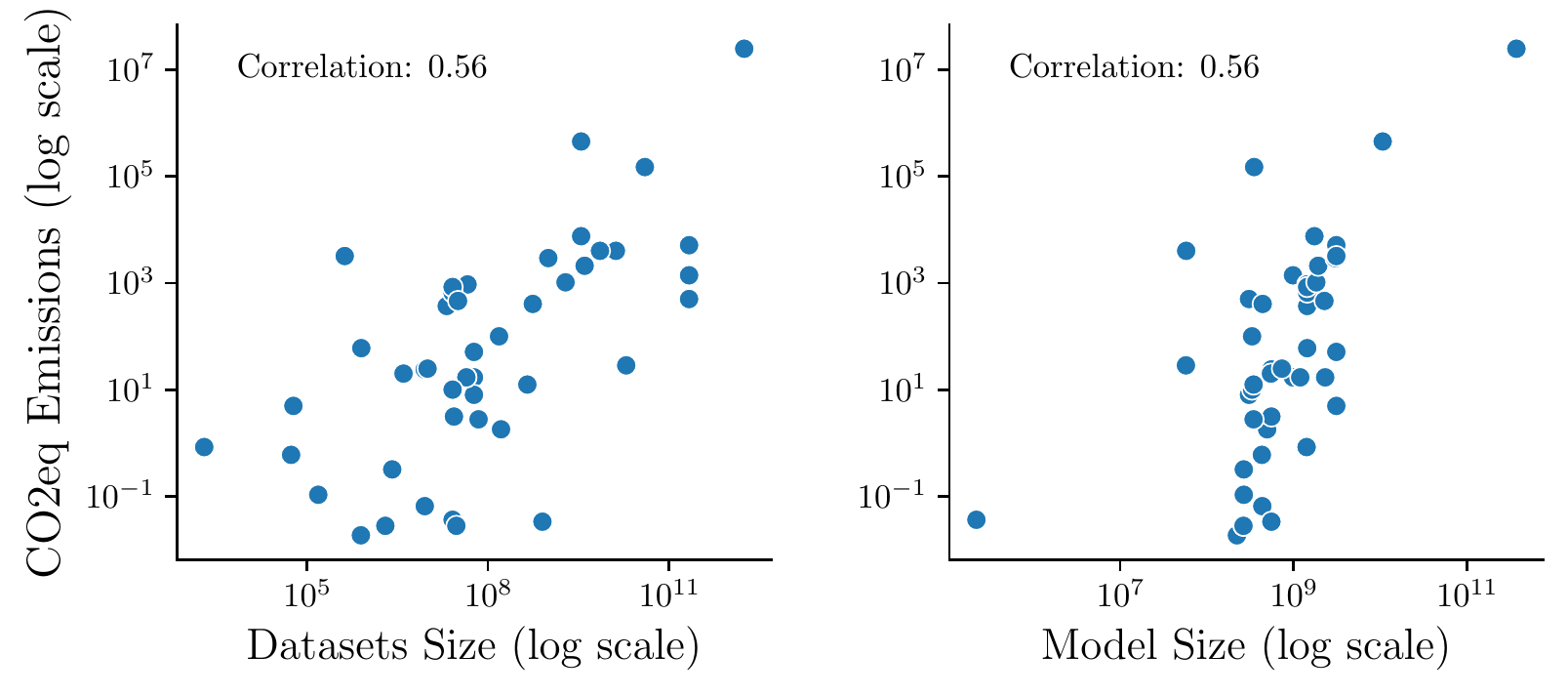}
    \caption{Correlation between carbon emissions and dataset / model size}
    \label{CarbonEmissionsBySize}
\end{figure}

Based on the correlation test between model size and carbon emissions with the curated dataset, we can conclude that there exists a positive correlation. Applying Pearson's correlation with the log-transformed variables to ensure normality, we obtain adjusted p-values $<0.05$ and correlations $\approx0.56$. These results are visualized in Figure~\ref{CarbonEmissionsBySize}.

\subsubsection{\textbf{What are the carbon emission differences between fine-tuned and pretrained tasks?}} 

~
\begin{myquote}
\textbf{Finding 2.3}. \textit{Although fine-tuning tasks appear to consume less than full pretraining tasks, we cannot conclude that the difference is statistically significant.}
\end{myquote}

\begin{figure}[h]
    \centering
    \includegraphics[width=0.85\columnwidth]{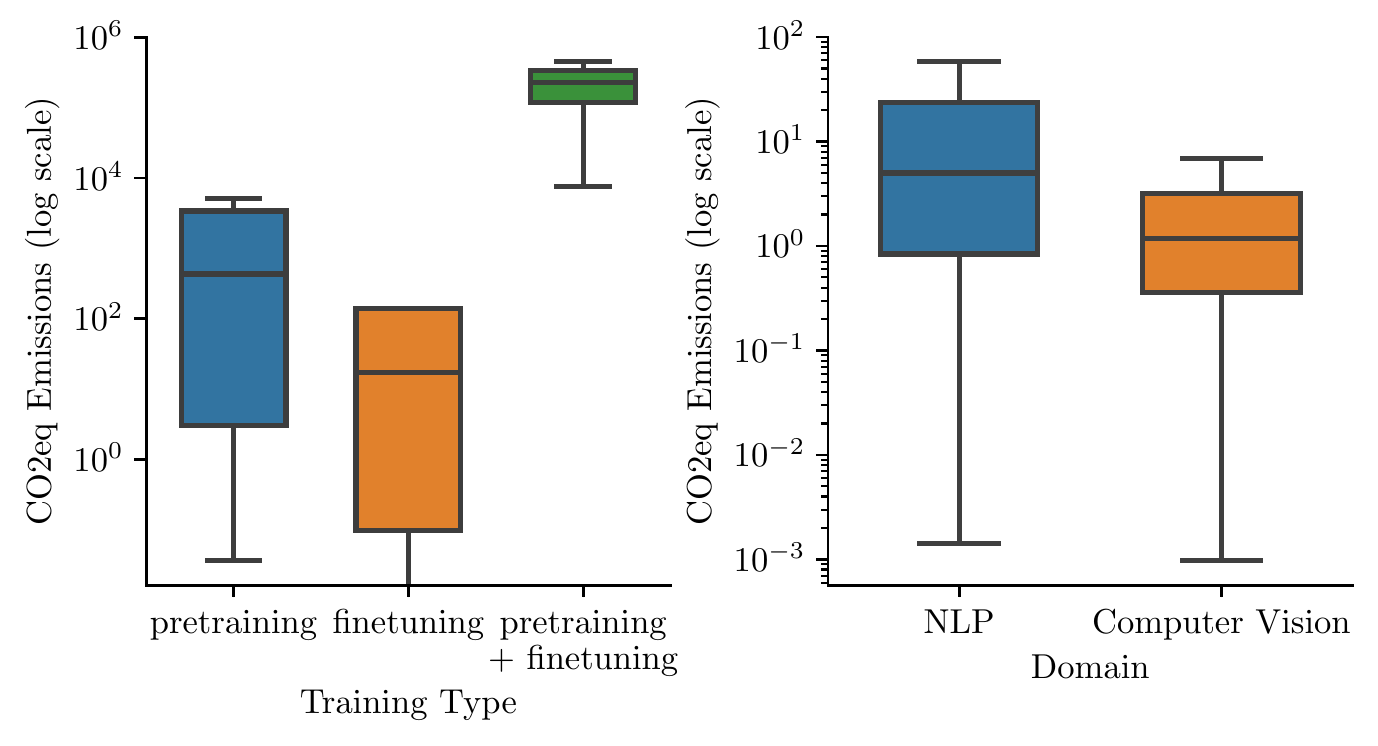}
    \caption{Carbon emissions by ML application domain and training type}
    \label{Domain&TrainingTypeEmissions}
\end{figure}

In Figure~\ref{Domain&TrainingTypeEmissions}, we can observe the carbon emissions divided by pretraining, fine-tuning, and pretraining + fine-tuning tasks. Pretraining involves training a model on a large dataset to learn general patterns and features, while fine-tuning refines the model on a specific task or domain using a smaller, targeted dataset. We can notice a difference between fine-tuning and pretraining tasks, where fine-tuning consumes on average 200\% less than pretraining.
Nonetheless, based on the Mann‐Whitney U test (adjusted p-value $=0.29$), we do not have enough evidence to conclude that this difference is, in fact, significant considering $\alpha=0.05$. Despite the visual indication of a difference in the plot, the adjusted p-value suggests that the observed difference might be due to chance.

We can also observe that tasks combining pretraining and fine-tuning consume more than the other two (e.g., pretraining median is 432.67g vs. 228.92kg on pretraining + fine-tuning). This is expected since integrating both training stages requires additional computational resources.

\subsubsection{\textbf{How do ML application domains affect carbon emissions?}}

~

\begin{myquote}
\textbf{Finding 2.4}. \textit{There is not enough evidence to consider that ML application domains affect carbon emissions.}
\end{myquote}

As we saw in RQ1.3, there is only a marginal number of audio and multimodal models that report carbon emissions. Thus, we exclude these ML application domains and only consider NLP and computer vision. According to Figure~\ref{Domain&TrainingTypeEmissions}, there is a slight difference in carbon emissions between NLP and vision models: the latter seem to consume less.
The Mann‐Whitney U test confirms (p-value = 1.14e-15) that there is a statistically significant difference in carbon emissions between the computer vision and NLP domains. However, this difference may be due to an increased model or dataset size for NLP models. In fact, NLP and model size are positively correlated (Spearman's  correlation coefficient = 0.41, p-value $\approx$ 0) and there is a significant (p-value = 2.58e-51) difference between model size for NLP and computer vision. Thus, we conclude the domain difference to be a confounding effect.


\subsubsection{\textbf{How can we classify Hugging Face models based on their carbon efficiency?}}

\begin{figure}[h]
    \centering
    \includegraphics[width=0.825\columnwidth]{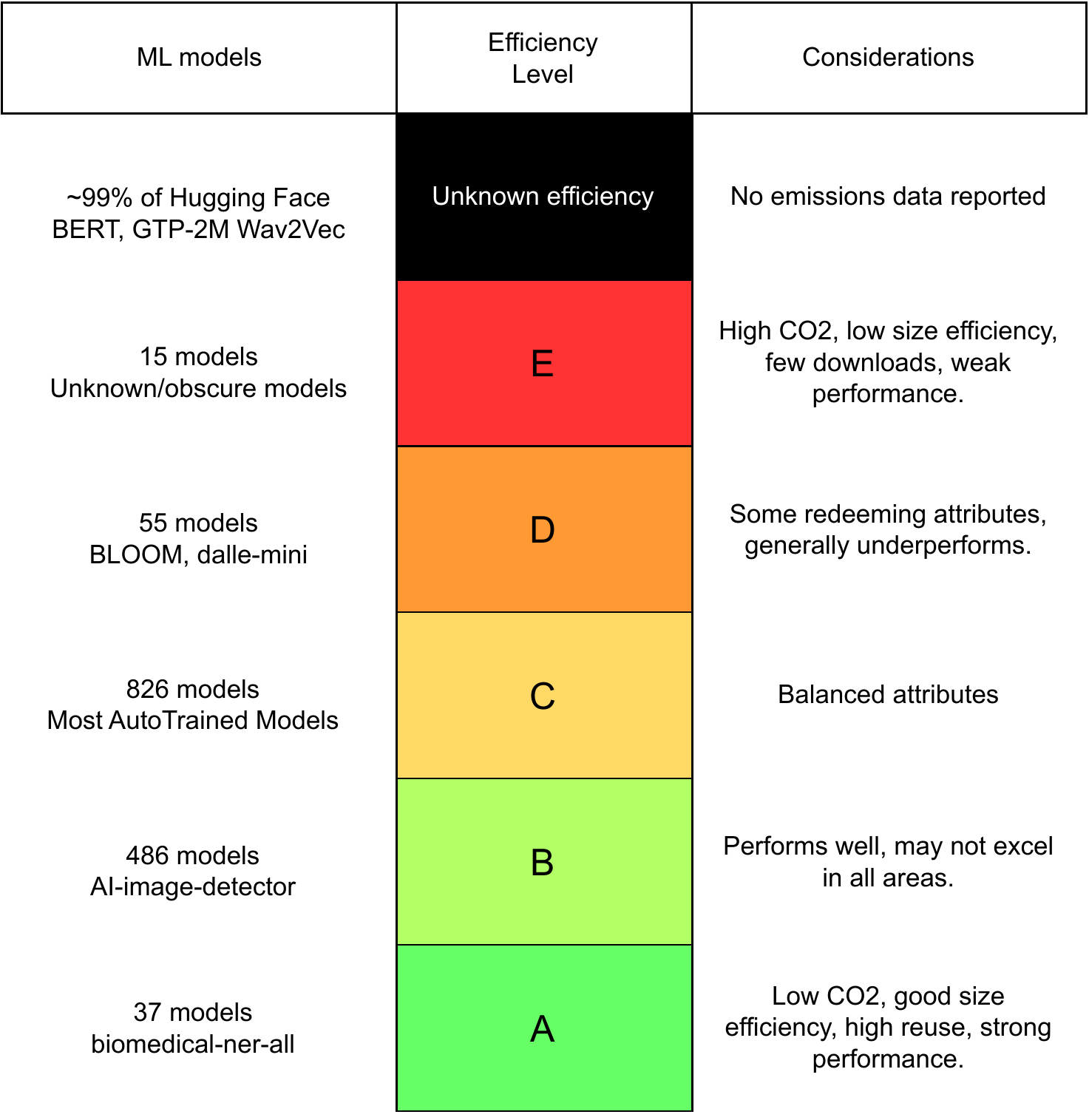}
    \caption{Carbon efficiency classification}
    \label{energy_efficiency_classification}
\end{figure}

Using the developed classification system, we categorize the models based on their carbon efficiency (see Figure~\ref{energy_efficiency_classification}).

We found the following distribution of Hugging Face models across carbon efficiency labels:

\begin{enumerate}
	\item E Label (15 models): These models have high  CO$_2$e emissions, relatively low model size, few downloads, and overall poor performance. Popular models are absent from this category due to low download counts.
	\item D Label (55 models): Similar to E Label, but with at least one attribute outperforming the others, e.g., \texttt{BLOOM} has high CO$_2$e (4.01kg), poor size efficiency, but much more downloads (47,449) than average E models. \texttt{distilpgt2}, our reference model, also classifies here.
	\item C Label (826 models): These models exhibit a balance between the attributes. Most of the AutoTrained models, which exhibit the average behaviour, fall here.
	\item B Label (486 models): Examples include \texttt{AI-image-detector}, which has 1K downloads, rather low emissions (7.94g) and good performance (0.94 accuracy), reflecting better carbon efficiency.
	\item A Label (37 models): These models demonstrate low CO$_2$e, high download counts, proper model size efficiency and good performance. Examples include: \texttt{biomedical-ner-all} with high reusability (15.5k downloads), low emissions (0.028g) and high complexity considering the low emissions; \texttt{BERT-Banking77} with high reusability (5,043 downloads), low emissions (0.033g) and good accuracy (0.926).
\end{enumerate}

This classification system based on weighted means provides valuable insights into the carbon efficiency profiles of Hugging Face models. This information can guide ML practitioners and researchers in selecting models that not only meet their specific requirements, but also contribute to more sustainable ML practices.

\section{Implications}

The findings of this paper have several important implications for the ML research community and industry practitioners. By identifying the current state of carbon emission reporting for ML models on Hugging Face, this research highlights areas of improvement and offers recommendations for promoting carbon-efficient ML development.

\subsection{Raising Awareness, Encouraging Transparency, and Standardizing Reporting Practices}

Our analysis reveals that despite the growing popularity of Hugging Face, the proportion of models reporting carbon emissions has stalled, suggesting a gap in ML sustainability awareness. Moreover, there is a clear lack of standardization in carbon emission reporting on Hugging Face, with ML practitioners reporting carbon emissions and emission context attributes without proper guidelines, leading to missing attributes (such as location, hardware, or model and datasets sizes). This lack of awareness and standardization in reporting practices may also be contributing to the absence of models meeting existing carbon efficiency classifications (e.g., \cite{fischer2023unified}), as we saw in RQ1.5 with the carbon reporting classification. 

Emphasizing the importance of reporting energy consumption can help promote standardized reporting guidelines and best practices for sustainable ML development. The AI community should actively promote an energy-efficient model development, transparent energy reporting practices, and energy efficiency certification \cite{8880037}. Establishing standardized reporting guidelines can help improve the consistency and quality of energy data and context reported.

To address these issues in carbon emissions reporting on Hugging Face, we propose the following initial guidelines:

\begin{itemize}
\item Carbon emissions (CO$_2$e): It should be reported consistently across all models to enable comparisons (gCO$_2$e). Some tools to track emissions are \texttt{CodeCarbon}~\cite{CodeCarbon}, \texttt{Carbontracker}~\cite{anthony2020carbontracker} or \texttt{Eco2AI}~\cite{budennyy2022eco2ai}.
\item Energy consumption metrics: Report energy consumption in kilowatt-hours (kWh) during training and inference, e.g., via tools such as NVML \cite{nvidiaNVIDIAManagement} or RAPL \cite{intelRunningAverage} and training time. This metric depends more on software design decisions and is more precise to judge efficiency than carbon emissions alone, which depends on carbon intensity. Further proposals on the estimation of energy consumption can be seen in \cite{Garcia-Martin2019, Cruz2022}.
\item Key energy-related context information: Include accurately attributes such as: hardware, location, energy source, model size, dataset size, and performance metrics. These factors can significantly impact the carbon footprint of a model and can be useful to evaluate the trade-off on carbon emissions.
\item Energy optimization techniques: Encourage reporting of any energy optimization techniques used during the model training and deployment process in the Model Card text. This information can help others in adopting similar practices to improve energy efficiency.
\end{itemize}

These guidelines could be translated into the following extended Hugging Face metadata proposed in \cite{huggingfaceDisplayingCarbon}:

\begin{lstlisting}[style=yaml]
co2_eq_emissions:
    emissions: number (in grams of CO2e).
    energy_consumption: number (in kWh).
    emissions_source: source of the information, e.g., code carbon, from AutoTrain, mlco2 calculator, etc.
    training_type: pre-training or fine-tuning.
    geographical_location: as granular as possible.
    hardware_used: how much compute and what kind, e.g., 8 v100 GPUs.
    cloud_service: cloud service name, if used.
    training_time: training duration (in seconds).
    optimization_techniques: any energy optimization techniques used during the model training and deployment process.
    energy_label: 
        - energy_label_source: certification label name and source.
          energy_label_classification: energy classification.
model_info:
    model_file_size: size of the model resulting file.
    number_of_parameters: number of parameters of the model.
    datasets_size: size of the dataset used.
    performance_metrics:
        - metric: e.g., accuracy.
          value: value of the metric, e.g., 0.92.
        - metric: e.g., f1.
          value: value of the metric.
\end{lstlisting}



\subsection{Developing Energy-Efficient Models and Enhancing Carbon Efficiency Classification}

Our proposal for classifying the carbon efficiency of ML models is based on these practices:

\begin{itemize}
    \item Minimize CO$_2$e emissions: Focus on reducing the environmental impact during model training and deployment.
    \item Encourage reusability: Share and promote models that can be easily adapted for various tasks, increasing their download count and overall efficiency. This aligns with RQ2.3 finding, as we could not find evidence to suggest that fine-tuning tasks are less consuming.
    \item Find the proper trade-off on model size and dataset size: Our findings show that larger models and datasets lead to increased carbon emissions during training.
    \item Maintain strong performance: Ensure models have balanced performance across key evaluation metrics. Our research could not find enough evidence to relate model performance with carbon emissions. Therefore, the expected trade-off may be more complex than expected. 
\end{itemize}

By adopting the above best practices and enhancing the carbon efficiency classification, researchers and developers can create models that meet specific requirement while contributing to environmentally sustainable ML practices.

\section{Threats to Validity}

Following empirical standards on repository mining \cite{repositoryMiningStandard}, we reported the applied good practices throughout the paper, e.g., explaining why Hugging Face mining is appropriate for our research goal, describing data pre-processing, and providing a detailed replication package. Below, we discuss several potential validity threats and associated mitigation actions.

\textbf{Construct Validity:} The reliance on self-reported carbon emission data from Hugging Face may not accurately reflect the actual carbon emissions due to variations in measurement methodologies or inconsistencies in reporting. Moreover, the manual curation of the carbon emission dataset could introduce bias in the selection and filtering of models. 

To address the self-reporting issue, mitigation actions were applied during the Data Preprocessing stage, which involved a harmonization process to standardize all model reports. Moreover, Hugging Face has implemented a carbon emission reporting proposal to address the self-reporting issue and AutoTrained models automatically report carbon emissions, which adds a layer of transparency and reduces the risk associated with self-reported data.

\textbf{Internal Validity:}
While we controlled for certain variables such as model size and datasets size, other factors like training setup, hardware, and data preprocessing might also influence the carbon emissions of a model. Additionally, the lack of informative model cards for many AutoTrained models might limit our ability to accurately assess the quality and representativeness of these models in our study.

\textbf{External Validity:}
Our findings based on Hugging Face models, a NLP predominant repository, may not be fully generalizable to other ML application domains like computer vision. Moreover, while our analysis covered a substantial sample of over 1,400 models (emissions-reporting models), it did not capture the entire range of models available in Hugging Face. Also, evolving trends in ML and energy efficiency could also affect the applicability of our findings in future contexts. These limitations can be mitigated through the replication package, allowing for data updates when desired.

\textbf{Conclusion Validity:}
The lack of standardized reporting practices for carbon emission data might have affected our ability to accurately compare models and draw conclusions. We made assumptions (included in the replication package \cite{castano_fernandez_joel_2023_8115187})  based on the available information, which may not always be correct or complete. To improve conclusion validity, future research should seek to establish more standardized reporting and data collection practices for carbon emissions.

\section{Conclusion}

This paper has addressed the crucial matter of carbon emissions and reporting practices in the ML community using the example of Hugging Face. Through a comprehensive analysis of carbon efficiency and reporting practices, we provided insights and recommendations for ML practitioners and researchers.
Despite the growing popularity of Hugging Face, we found a concerning lack of ML sustainability awareness, as illustrated by the stalled proportion of carbon emissions reporting models. Additionally, the lack of standardization in emissions reporting practices highlights the need for establishing guidelines and promoting energy efficiency certification.

By examining the correlations between carbon emissions and other factors, we identified important insights related to carbon efficiency in Hugging Face ML models. Moreover, we have proposed a carbon efficiency classification system that can guide ML practitioners in their model selection, contributing to more environmentally sustainable ML practices.

It is crucial for the ML community to work together to raise awareness, encourage transparency, standardize reporting practices, and develop energy-efficient models across various ML application domains.
The findings and recommendations presented in this paper serve as a stepping stone for future research and development in the field of Green AI. It is our hope that this work inspires continued efforts to create more sustainable and environmentally friendly ML systems, contributing to a greener, more responsible future for ML.

\section{ACKNOWLEDGMENT}
This work is partially supported by the project TED2021-130923B-I00, funded by MCIN/AEI/10.13039/501100011033 and the European Union Next Generation EU/PRTR.

\bibliographystyle{IEEEtran}
\bibliography{References}

\end{document}